\documentclass[conference, onecolumn]{IEEEtran}
\IEEEoverridecommandlockouts
\usepackage{amsthm,amsmath,amssymb,amsfonts,bm}
\usepackage{algorithmic}
\usepackage{graphicx}
\usepackage{textcomp}
\usepackage{xcolor}
\usepackage{subfigure}
\usepackage{booktabs}

\usepackage{multirow}
\usepackage{makecell}

\theoremstyle{plain}
\newtheorem{theorem}{Theorem}[section]

\newtheorem{theorem_app}{Theorem}[section]
\newtheorem{lemma_app}{Lemma}[section]

\theoremstyle{definition}
\newtheorem{axiom}{Axiom}
\newcommand{\Ll}{\mathcal{L}}
\newcommand{\X}{\mathcal{X}}
\newcommand{\Y}{\mathcal{Y}}
\newcommand{\Pp}{\mathbb{P}}
\newcommand{\E}{\mathbb{E}}
\newcommand{\x}{\bm{x}}
\newcommand{\y}{\bm{y}}
\newcommand{\z}{\bm{z}}
\newcommand{\TP}{\text{TP}}
\newcommand{\TN}{\text{TN}}
\newcommand{\FP}{\text{FP}}
\newcommand{\FN}{\text{FN}}

\usepackage{tikz}
\usetikzlibrary{shapes.geometric, arrows}
\tikzstyle{startstop} = [ellipse, minimum width=1.5cm, minimum height=0.7cm,text centered, draw=black, text width=1.5cm, font=\scriptsize]

\tikzstyle{io} = [trapezium, trapezium left angle=70, trapezium right angle=110, minimum width=0.5cm, minimum height=0.7cm, text centered, draw=black, text width=1.2cm, font=\scriptsize]
\tikzstyle{process} = [rectangle, minimum width=2cm, minimum height=0.7cm, text centered, draw=black, text width=2cm, font=\scriptsize]
\tikzstyle{decision} = [diamond, minimum width=1.5cm, minimum height=0.7cm, text centered, draw=black, text width=2.1cm, aspect=2, inner sep=-1pt, font=\scriptsize]

\tikzstyle{arrow} = [thick,->,>=stealth, draw=blue!50]

\usepackage[backend=biber,style=ieee,sorting=none]{biblatex}
\usepackage{url}
\def\BibTeX{{\rm B\kern-.05em{\sc i\kern-.025em b}\kern-.08em
    T\kern-.1667em\lower.7ex\hbox{E}\kern-.125emX}}
    
\defbibheading {bibliography}[References]{\section*{#1}}
\usepackage{hyperref}

\usepackage{mathtools}

\begin{document}

\title{An Asymmetric Contrastive Loss for Handling Imbalanced Datasets}

\author{
\begin{tabular}[t]{c@{\extracolsep{4em}}c} 
Valentino Vito & Lim Yohanes Stefanus \\
\textit{Faculty of Computer Science} & \textit{Faculty of Computer Science} \\
\textit{Universitas Indonesia} & \textit{Universitas Indonesia} \\
\textit{Depok 16424, Indonesia} & \textit{Depok 16424, Indonesia} \\
\textit{valentino.vito11@ui.ac.id} & \textit{yohanes@cs.ui.ac.id}
\end{tabular}
}

\maketitle

\begin{abstract}
Contrastive learning is a representation learning method performed by contrasting a sample to other similar samples so that they are brought closely together, forming clusters in the feature space. The learning process is typically conducted using a two-stage training architecture, and it utilizes the contrastive loss (CL) for its feature learning. Contrastive learning has been shown to be quite successful in handling imbalanced datasets, in which some classes are overrepresented while some others are underrepresented. However, previous studies have not specifically modified CL for imbalanced datasets. In this work, we introduce an asymmetric version of CL, referred to as ACL, in order to directly address the problem of class imbalance. In addition, we propose the asymmetric focal contrastive loss (AFCL) as a further generalization of both ACL and focal contrastive loss (FCL). Results on the FMNIST and ISIC 2018 imbalanced datasets show that AFCL is capable of outperforming CL and FCL in terms of both weighted and unweighted classification accuracies. In the appendix, we provide a full axiomatic treatment on entropy, along with complete proofs.
\end{abstract}

\begin{IEEEkeywords}
Asymmetric loss, class imbalance, contrastive loss, entropy, focal loss.
\end{IEEEkeywords}

\section{Introduction}
Class imbalance is a major obstacle occurring within a dataset when certain classes in the dataset are overrepresented (referred to as majority classes), while some are underrepresented (referred to as minority classes). This can be problematic for a large number of classification models. A deep learning model such as a convolutional neural network (CNN) might not be able to properly learn from the minority classes. Consequently, the model would be less likely to correctly identify minority samples as they occur. This is especially crucial in medical imaging, since a model that cannot identify rare diseases would not be effective for diagnostic purposes. For example, the ISIC 2018 dataset \cite{codella,tschandl} is an imbalanced medical dataset which consists of images of skin lesions that appear in various frequencies during screening.

To produce a less imbalanced dataset, it is possible to resample the dataset by either increasing the number of minority samples \cite{bej, fajardo, karia, tripathi} or decreasing the number of majority samples \cite{arefeen, dai, koziarski, rayhan}. Other methods to handle class imbalance include substituting the standard cross-entropy (CE) loss for a more suitable loss, such as the focal loss (FL). Lin \textit{et al.} \cite{lin_focal} modified the CE loss into FL so that minority classes can be prioritized. This is done by ensuring that the model focuses on samples that are harder to classify during model training. Recent studies also unveiled the potential of contrastive learning as a way to combat imbalanced datasets \cite{marrakchi, chen_supercon}.

Contrastive learning is performed by contrasting a sample (called an \textit{anchor}) to other similar samples (called positive samples) so that they are mapped closely together in the feature space. As a consequence, dissimilar samples (called negative samples) are pushed away from the anchor, forming clusters in the feature space based on similarity. In this research, contrastive learning is done using a two-stage training architecture, which utilizes the contrastive loss (CL) formulated by Khosla \textit{et al.} \cite{khosla}. This formulation of CL is supervised based, and it can contrast the anchor to multiple positive samples belonging to the same class. This is unlike self-supervised contrastive learning \cite{chen_simple, henaff, hjelm, tian}, which contrasts the anchor to only one positive sample in the mini-batch.

In this work, we propose a modification of CL, referred to as the asymmetric contrastive loss (ACL). Unlike CL, the ACL is able to directly contrast the anchor to its negative samples so that they are pushed apart in the feature space. This becomes important when a rare sample has no other positive samples in the mini-batch. To our knowledge, this is the first study to modify CL directly in order to address the class imbalance problem. We also consider the asymmetric variant of the focal contrastive loss (FCL) \cite{zhang_fcl}, called the asymmetric focal contrastive loss (AFCL). Using FMNIST and ISIC 2018 as datasets, experiments are done to test the performance of both ACL and AFCL in binary classification tasks. It is observed that AFCL is superior to CL and FCL in multiple class-imbalance scenarios, provided that suitable hyperparameters are used. In addition, this work provides a streamlined survey on the literature related to entropy and loss functions.

\section{Background on Entropy and Loss Functions}
In this section, we provide a literature review on basic information theory and various loss functions.

\subsection{Entropy, Information, and Divergence}
Introduced by Shannon \cite{shannon}, entropy provides a measure on the amount of information contained in a random variable, usually in bits. The \textit{entropy} $H(X)$ of a random variable $X$ is given by the formula
\begin{equation}
    H(X) = \E_{P_X}\left[-\log(P_X(X))\right].
\end{equation}
Given two random variables $X$ and $Y$, their \textit{joint entropy} $H(X, Y)$ is the entropy of the joint random variable $(X,Y)$:
\begin{equation}
    H(X,Y) = \E_{P_{(X,Y)}}\left[-\log(P_{(X,Y)}(X,Y))\right].
\end{equation}
In addition, the \textit{conditional entropy} $H(Y \mid X)$ is defined as
\begin{equation}
    H(Y \mid X) = \E_{P_{(Y,X)}}\left[-\log(P_{Y \mid X}(Y \mid X)\right].
\end{equation}
Conditional entropy is used to measure the average amount of information contained in $Y$ when the value of $X$ is given. Conditional entropy is bounded above by the original entropy; that is, $H(Y \mid X) \le H(Y)$, with equality if and only if $X$ and $Y$ are independent \cite{ajjanagadde}.

The formulas for entropy, joint entropy, and conditional entropy can be derived via an axiomatic approach \cite{gowers, khinchin}. The list of axioms is provided in Appendix \ref{app:entropy}, whereas the derivation of the formula of entropy is provided in Appendix \ref{app:entropy_proof}.

The \textit{mutual information} $I(X;Y)$ is a measure of dependence between random variables $X$ and $Y$ \cite{cover}. It provides the amount of information about one random variable provided by the other random variable, and it is defined by
\begin{equation}
    I(X;Y) = H(X) - H(X \mid Y) = H(Y) - H(Y \mid X).
\end{equation}
Mutual information is symmetric. In other words, $I(X;Y) = I(Y;X)$. Mutual information is also nonnegative ($I(X;Y) \ge 0$), and $I(X;Y) = 0$ if and only if $X$ and $Y$ are independent \cite{ajjanagadde}.

The dissimilarity between random variables $X$ and $X'$ on the same space $\X$ can be measured using the notion of \textit{KL-divergence}:
\begin{equation}
    D_{\text{KL}}(X \,\Vert\, X') = \E_{P_X}\left[\log\left(\frac{P_X(X)}{P_{X'}(X)}\right)\right].
\end{equation}
Similar to mutual information, KL-divergence is nonnegative ($D_{\text{KL}}(X \,\Vert\, X') \ge 0$), and $D_{\text{KL}}(X \,\Vert\, X') = 0$ if and only if $X = X'$ \cite{ajjanagadde}. Unlike mutual information, KL-divergence is asymmetric, so $D_{\text{KL}}(X \,\Vert\, X')$ and $D_{\text{KL}}(X' \,\Vert\, X)$ are not necessarily equal.

\subsection{Cross-Entropy and Focal Loss}
Given random variables $X$ and $\hat{X}$ on the same space $\X$, their \textit{cross-entropy} $H(X;\hat{X})$ is defined as \cite{boudiaf}:
\begin{equation}\label{eq:cross-entropy}
    H(X;\hat{X}) = \E_{P_X}\left[-\log(P_{\hat{X}}(X)\right].
\end{equation}
Cross-entropy is the average amount of bits needed to encode the true distribution $X$ when its estimate $\hat{X}$ is provided \cite{murphy}. A small value of $H(X;\hat{X})$ implies that $\hat{X}$ is a good estimate for $X$. Cross-entropy is connected to KL-divergence via the following identity:
\begin{equation}
    H(X;\hat{X}) = H(X) + D_{\text{KL}}(X \,\Vert\, \hat{X}).
\end{equation}
When $\hat{X} = X$, the equality $H(X;\hat{X}) = H(X)$ holds.

Now, the cross-entropy loss and focal loss are provided within the context of a binary classification task consisting of two classes labeled $0$ and $1$. Suppose that $y \in \{0, 1\}$ denotes the ground-truth class and $p \in [0, 1]$ denotes the estimated probability for the class labeled $1$. The value of $1-p$ is then the estimated probability for the class labeled $0$. The \textit{cross-entropy (CE) loss} is given by
\begin{align*}
    \Ll_{\text{CE}} &= - y \log(p) - (1-y) \log(1-p)\\
    &=
    \begin{cases}
        -\log(p)  & y = 1, \\
        -\log(1-p) & y = 0.
    \end{cases}
\end{align*}

If $y = 1$, then the loss $\Ll_{\text{CE}}$ is zero when $p = 1$. On the other hand, if $y = 0$, then the loss is zero when $1 - p = 1$. In either case, the CE loss is minimized when the estimated probability of the true class is maximized, which is the desired property of a good classification model.

The \textit{focal loss} (FL) \cite{lin_focal} is a modification of the CE loss introduced to put more focus on hard-to-classify examples. It is given by the following formula:
\begin{equation}
    \Ll_{\text{foc}} = - y (1-p)^\gamma \log(p) - (1-y) p^\gamma \log(1-p).
\end{equation}

\begin{figure}[t]
	\centering
	\includegraphics[width=0.6\textwidth]{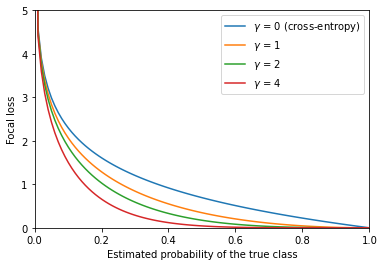}
	\caption{A graph illustrating the focal loss given the predicted probability of the ground-truth class, with varying values of $\gamma$}
	\label{fig:focal-loss}
\end{figure}

The parameter $\gamma$ in $\Ll_{\text{foc}}$ is known as the \textit{focusing parameter}. Choosing a larger value of $\gamma$ would push the model to focus on training from the misclassified examples. For instance, suppose that $\gamma = 4$ and denote the estimated probability of the true class by $p_t$. The graph on Figure \ref{fig:focal-loss} shows that when $p_t > 0.5$, the FL is quite small. Hence, the model would be less concerned about learning from an example when $p_t$ is already sufficiently large. FL is a useful choice when class imbalance exists as it can help the model focus on the less represented samples within the dataset.

\subsection{Asymmetric Loss}
For multi-label classification with $K$ labels, let $y_i \in \{0, 1\}$ be the ground truth for class $i$ and $p_i \in [0, 1]$ be its estimated probability obtained by the model. The aggregate classification loss is then
\begin{equation}
    \Ll = \sum_{i=1}^K \Ll_i,
\end{equation}
where
\begin{equation}\label{eq:multilabel}
    \Ll_i = - y_i \Ll_i^+ - (1-y_i) \Ll_i^-.
\end{equation}
If FL is the chosen type of loss, $\Ll_i^+$ and $\Ll_i^-$ are set as follows:
\begin{equation}\label{eq:multilabel1}
    \Ll_i^+ = (1-p_i)^\gamma\log(p_i) \quad \text{and} \quad \Ll_i^- = p_i^\gamma\log(1-p_i).
\end{equation}

In a typical multi-label dataset, the ground truth $y_i$ has value $0$ for the majority of classes $i$. Consequently, the negative terms $\Ll_i^-$ dominate in the calculation of the aggregate loss $\Ll$. \textit{Asymmetric loss} (ASL) \cite{ben-baruch} is a proposed solution to this problem. ASL emphasizes the contribution of the positive terms by modifying the losses of Eq.\ (\ref{eq:multilabel1}) to
\begin{equation}
    \Ll_i^+ = (1-p_i)^{\gamma^+}\log(p_i)
\end{equation}
and
\begin{equation}
    \Ll_i^- = (p_i^{(m)})^{\gamma^-}\log(1-p_i^{(m)}),
\end{equation}
where $\gamma^+, \gamma^-$ are hyperparameters and $p_i^{(m)}$ is the \textit{shifted probability} of $p_i$ obtained from the \textit{probability margin} $m \ge 0$ via the formula
\begin{equation}
    p_i^{(m)} = \max(p_i-m, 0).
\end{equation}
This shift helps decrease the contribution of $\Ll_i^-$. Indeed, if we set $m = 1$, then $\Ll_i^- = 0$.

\subsection{Contrastive Loss}
Contrastive learning is a learning method to learn representations from data. A supervised approach of contrastive learning was introduced by Khosla et al.\ \cite{khosla} to learn from a set of sample-label pairs $\{(\x_i, \y_i)\}_{i=1}^N$ in a mini-batch of size $N$. The samples $\x_i$ are fed through a feature encoder $\text{Enc}(\cdot)$ and a projection head $\text{Proj}(\cdot)$ in succession to obtain features $\z_i = \text{Proj}(\text{Enc}(\x_i))$. The feature encoder extracts features from $\x_i$, whereas the projection head projects the features into a lower dimension and apply $\ell_2$-normalization so that $\z_i$ lies in the unit hypersphere. In other words, $\lVert \z_i \rVert_2 = 1$.

A pair $(\z_i,\z_j)$, where $i \neq j$, is referred to as a \textit{positive pair} if the features share the same class label ($\y_i = \y_j$) and it is a \textit{negative pair} if the features have different class labels ($\y_i \neq \y_j$). Contrastive learning aims to maximize the similarity between $\z_i$ and $\z_j$ whenever they form a positive pair and minimize their similarity whenever they form a negative pair. This similarity is measured with cosine similarity \cite{murphy}:
\begin{equation}
    \kappa(\z_i, \z_j) = \frac{\z_i \cdot \z_j}{\lVert \z_i \rVert_2 \lVert \z_j \rVert_2} = \z_i \cdot \z_j.
\end{equation}
From the above equation, we have $\kappa(\z_i, \z_j) \in [-1,1]$. In addition, $\kappa(\z_i, \z_j) = 1$ when $\z_i = \z_j$, and $\kappa(\z_i, \z_j) = -1$ when $\z_i$ and $\z_j$ form a $180^\circ$ angle.

Fixing $\z_i$ as the anchor, let $A_i = \{\z_k \mid k \neq i\}$ be the set of features other than $\z_i$ and let $P_i = \{\z_k \in A_i \mid \y_k = \y_i\}$ be the set of $\z_k$ such that $(\z_i, \z_k)$ is a positive pair. The predicted probability $p_{ij}$ that $\z_i$ and $\z_j$ belong to the same class is obtained by applying the softmax function to the the set of similarities between $\z_i$ and $\z_k \in A_i$:
\begin{equation}\label{eq:p_{ij}}
    p_{ij} = \frac{\exp(\z_i \cdot \z_j/\tau)}{\sum_{\z_k \in A_i} \exp(\z_i \cdot \z_k/\tau)},
\end{equation}
where $\tau$ is referred to as the \textit{temperature parameter}. Since our goal is to maximize $p_{ij}$ whenever $\z_j \in P_i$, the \textit{contrastive loss} which is to be minimized is formulated as
\begin{equation}\label{eq:contrastive}
    \Ll_{\text{con}} = -\sum_{i=1}^n \frac{1}{|P_i|} \sum_{\z_j \in P_i} \log(p_{ij}).
\end{equation}

Information-theoretical properties of $\Ll_{\text{con}}$ are given in \cite{zhang_fcl}, from which we provide a summary. Let $X$, $Y$, and $Z$ denote random variables of the samples, labels, and features, respectively. The following theorem states that $\Ll_{\text{con}}$ is positive proportional to $H(Z \mid Y) - H(Z)$ under the assumption that no class imbalance exists.

\begin{theorem}[Zhang et al.\ \cite{zhang_fcl}]\label{thm:contrastive_1}
Assuming that features are $\ell_2$-normalized and the dataset is balanced,
\begin{equation}
    \Ll_{\text{con}} \propto H(Z \mid Y) - H(Z).
\end{equation}
\end{theorem}

Theorem \ref{thm:contrastive_1} implies that minimizing $\Ll_{\text{con}}$ is equivalent to minimizing the conditional entropy $H(Z \mid Y)$ and maximizing the feature entropy $H(Z)$. Since $I(Z;Y) = H(Z)  - H(Z \mid Y)$, minimizing $\Ll_{\text{con}}$ is equivalent to maximizing the mutual information $I(Z;Y)$ between features $Z$ and class labels $Y$. In other words, contrastive learning aims to extract the maximum amount of information from class labels and encode them in the form of features.

After the features are extracted, a classifier $\text{Clas}(\cdot)$ is assigned to convert $\z_i$ into a prediction $\hat{\y}_i = \text{Clas}(\z_i)$ of the class label. The random variable of predicted class labels is denoted by $\hat{Y}$.

For the next theorem, the definition of \textit{conditional cross-entropy} $H(Y;\hat{Y} \mid Z)$ is given as follows:
\begin{equation}\label{eq:conditional-ce}
    H(Y;\hat{Y} \mid Z) = \E_{P_{(Y,Z)}}\left[-\log(P_{(\hat{Y},Z)}(Y,Z)\right].
\end{equation}
Conditional CE measures the average amount of information needed to encode the true distribution $Y$ using its estimate $\hat{Y}$, given the value of $Z$. A small value of $H(Y;\hat{Y} \mid Z)$ implies that $\hat{Y}$ is a good estimate for $Y$, given $Z$.

\begin{theorem}[Zhang et al.\ \cite{zhang_fcl}]\label{thm:contrastive_2}
Assuming that features are $\ell_2$-normalized and the dataset is balanced,
\begin{equation}
    \Ll_{\text{con}} \propto \inf H(Y;\hat{Y} \mid Z) - H(Y),
\end{equation}
where the infimum is taken over classifiers.
\end{theorem}

Theorem \ref{thm:contrastive_2} implies that minimizing $\Ll_{\text{con}}$ will minimize the infimum of conditional cross-entropy $H(Y;\hat{Y} \mid Z)$ taken over classifiers. As a consequence, contrastive learning is able to encode features in $Z$ such that the best classifier can produce a good estimate of $Y$ given the information provided by the feature encoder.

The formula for $\Ll_{\text{con}}$ can be modified so as to resemble the focal loss, resulting in a loss function known as the \textit{focal contrastive loss} (FCL) \cite{zhang_fcl}:

\begin{equation}\label{eq:fcl}
    \Ll_{\text{FC}} = -\sum_{i=1}^n \frac{1}{|P_i|} \sum_{\z_j \in P_i} (1 - p_{ij})\log(p_{ij}).
\end{equation}

\section{Methodology}
In this section, our proposed modification of the contrastive loss, called the asymmetric contrastive loss, is introduced. Also, the architecture of the model in which the contrastive losses are implemented is explained.

\subsection{Asymmetric Contrastive Loss}\label{sec:acl}
In Eq.\ (\ref{eq:contrastive}), the inside summation of the contrastive loss is evaluated over $P_i$. Consequently, according to Eq.\ (\ref{eq:p_{ij}}), each anchor $\z_i$ is contrasted with vectors $\z_j$ that belong to the same class. This does not present a problem when the mini-batch contains plenty of examples from each class. However, the calculated loss may not give each class a fair contribution when some classes are less represented in the mini-batch.

\begin{figure}[t]
	\centering
	\includegraphics[width=0.3\textwidth]{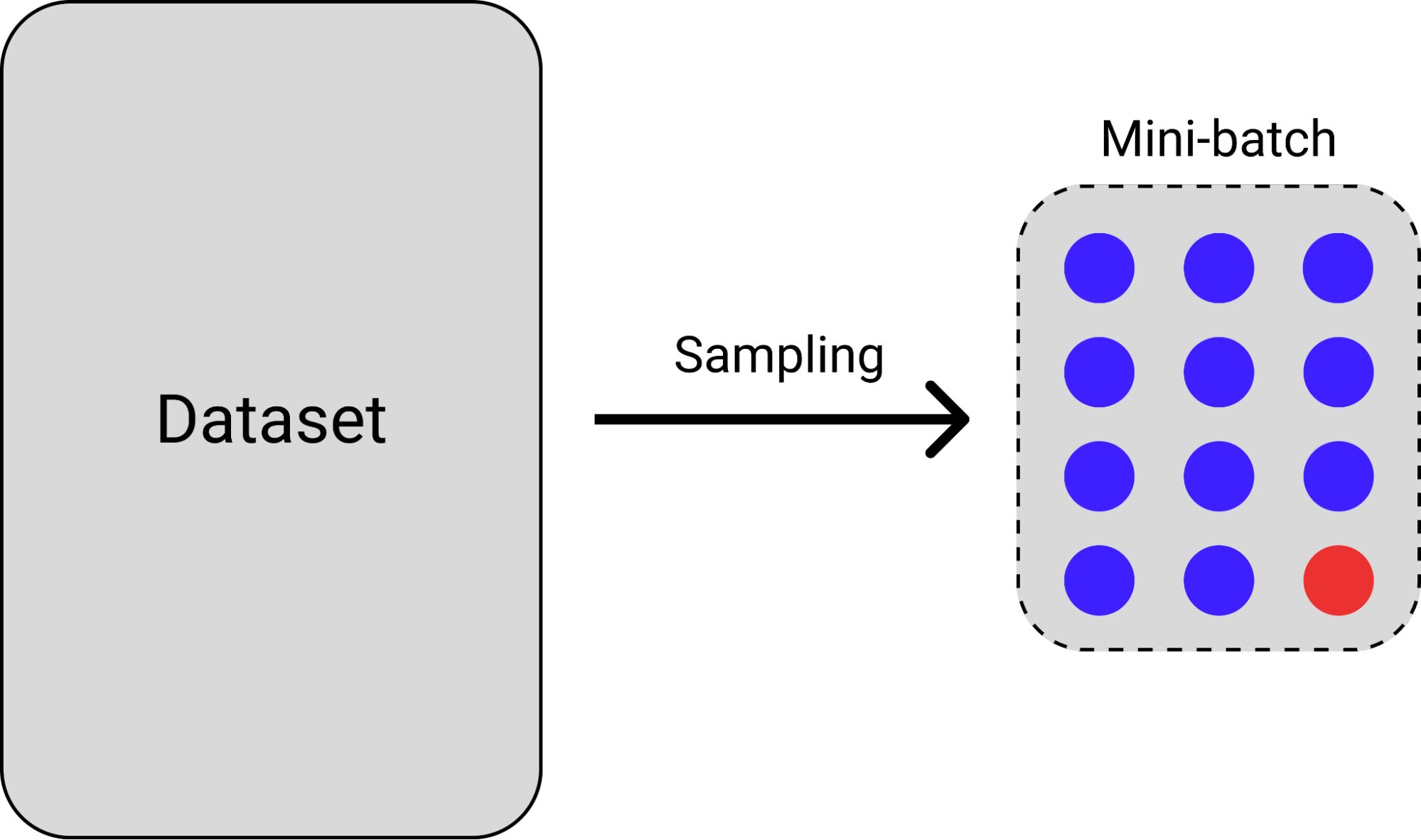}
	\caption{A mini-batch consisting of $11$ examples with blue-colored class label and $1$ example with red-colored class label}
	\label{fig:sampling}
\end{figure}

In Figure \ref{fig:sampling}, a sampled mini-batch consists of $11$ examples with blue-colored class label and $1$ example with red-colored class label. When the anchor $\z_i$ is the representation of the red-colored sample, $\z_i$ does not directly contribute to the calculation of $\Ll_{\text{con}}$ since $P_i$ is empty. In other words, $\z_i$ cannot be contrasted to any other sample in the mini-batch. This scenario is likely to happen when the dataset is imbalanced, and it motivates us to modify CL so that each anchor $\z_i$ can also be contrasted with $\z_j$ not belonging to the same class.

Let $N_i = A_i \setminus P_i$ be the set of vectors $\z_k$ such that $(\z_i, \z_k)$ is a negative pair. Motivated by the $\Ll_i^+$ and $\Ll_i^-$ of Eq.\ (\ref{eq:multilabel}), we define
\begin{equation}
    \Ll_i^+ = \frac{1}{|P_i|} \sum_{\z_j \in P_i} \log(p_{ij})
\end{equation}
and
\begin{equation}
    \Ll_i^- = \frac{1}{|N_i|} \sum_{\z_j \in N_i} \log(1-p_{ij}),
\end{equation}
where $p_{ij} = \exp(\z_i \cdot \z_j/\tau)/\sum_{\z_k \in A_i} \exp(\z_i \cdot \z_k/\tau)$. The loss function $\Ll_i^+$ contrasts $\z_i$ to vectors in $P_i$, whereas $\Ll_i^-$ contrasts $\z_i$ to vectors in $N_i$. The resulting \textit{asymmetric contrastive loss} (ACL) is given by the formula
\begin{equation}
    \Ll_{\text{AC}} = -\sum_{i=1}^n (\Ll_i^+ + \eta\Ll_i^-),
\end{equation}
where $\eta \ge 0$ is a fixed hyperparameter. If $\eta = 0$, then $\Ll_{\text{AC}} = \Ll_{\text{con}}$. Hence ACL is a generalization of CL.

When the batch size is set to a large number (over 100, for example), the value $p_{ij}$ tends to be very small. This causes $\Ll_i^-$ to be much smaller than $\Ll_i^+$. In order to balance their contribution to the total loss $\Ll_{\text{AC}}$, a large value for $\eta$ is usually chosen (between 60 and 300 in our experiment).

\subsection{Asymmetric Focal Contrastive Loss}
Following the formulation of $\Ll_{\text{FC}}$ in Eq.\ (\ref{eq:fcl}), $\Ll_i^+$ can be modified to have the following formula:
\begin{equation}
    \Ll_i^+ = \frac{1}{|P_i|} \sum_{\z_j \in P_i} (1-p_{ij})^{\gamma} \log(p_{ij}).
\end{equation}
Using this loss, the \textit{asymmetric focal contrastive loss} (AFCL) is then given by
\begin{equation}
    \Ll_{\text{AFC}} = -\sum_{i=1}^n (\Ll_i^+ + \eta\Ll_i^-),
\end{equation}
where $\Ll_i^- = \frac{1}{|N_i|} \sum_{\z_j \in N_i} \log(1-p_{ij})$. We do not modify $\Ll_i^-$ by adding the multiplicative term $(p_{ij})^{\gamma}$ since $p_{ij}$ is usually too small and would make $\Ll_i^-$ vanish if the term is added.

We have $\Ll_{\text{AFC}} = \Ll_{\text{FC}}$ when $\gamma = 1$. Thus, AFCL generalizes FCL. Unlike FCL, we add the hyperparameter $\gamma \ge 0$ to the loss function so as to provide some flexibility to the loss function.

\subsection{Model Architecture}\label{sec:architecture}
\begin{figure}[t]
	\centering
	\includegraphics[width=0.8\textwidth]{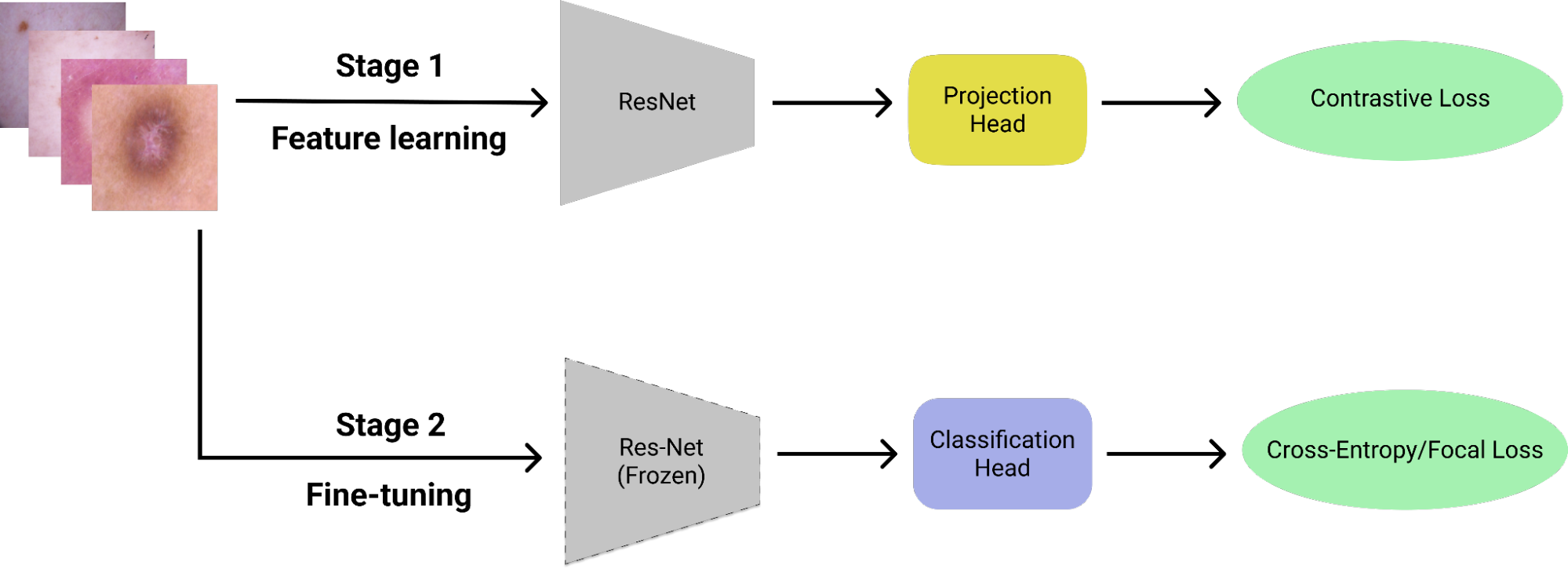}
	\caption{A two-stage training strategy consisting of: (1) feature learning using contrastive loss, and (2) classifier fine-tuning using either FL or CE loss}
	\label{fig:architecture}
\end{figure}

This section explains the inner workings of the classification model used for the implementation of the contrastive losses. The architecture of the model is taken from \cite{marrakchi,chen_supercon}. The training strategy for the model, as shown in Figure \ref{fig:architecture}, comprises of two stages: the feature learning stage and the fine-tuning stage.

In the first stage, each mini-batch is fed through a feature encoder. We consider either ResNet-18 or ResNet-50 \cite{he_resnet} for the architecture of the feature encoder. The output of the feature encoder is projected by the projection head to generate a vector $\z$ of length $128$. If ResNet-18 is used for the feature encoder, then the projection head consists of two layers of length 512 and 128. If ResNet-50 is used, then the two layers are of length 2048 and 128. Afterwards, $\z$ is $\ell_2$-normalized and the model parameters are updated using some version of the contrastive loss (either CL, FCL, ACL, or AFCL).

After the first stage is complete, the feature encoder is frozen and the projection head is removed. In its place, we have a one-layer classification head which generates the estimated probability that the training sample belongs to a certain class. The parameters of the classification head are updated using either the FL or CE loss. The final classification model is the feature encoder trained during the first stage, together with the classification head trained during the second stage. Since the classification head is a significantly smaller architecture than the feature encoder, training is mostly focused on the first stage. As a consequence, we typically need a larger number of epochs for the feature learning stage compared to the fine-tuning stage.

\section{Experiments}
The datasets and settings of our experiments are outlined in this section. We provide and discuss the results of the experiments on the FMNIST and ISIC 2018 datasets. The PyTorch implementation is available on GitHub \footnote{\url{https://github.com/valentinovito/Asymmetric-CL}}.

\subsection{Datasets}
In our experiments, the training strategy outlined in Subsection \ref{sec:architecture} is applied to two imbalanced datasets. The first is a modified version of the Fashion-MNIST (FMNIST) dataset \cite{xiao_fmnist}, and the second is the International Skin Imaging Collaboration (ISIC) 2018 medical dataset \cite{codella,tschandl}.

\begin{figure}[t]
	\centering
	\includegraphics[width=0.6\textwidth]{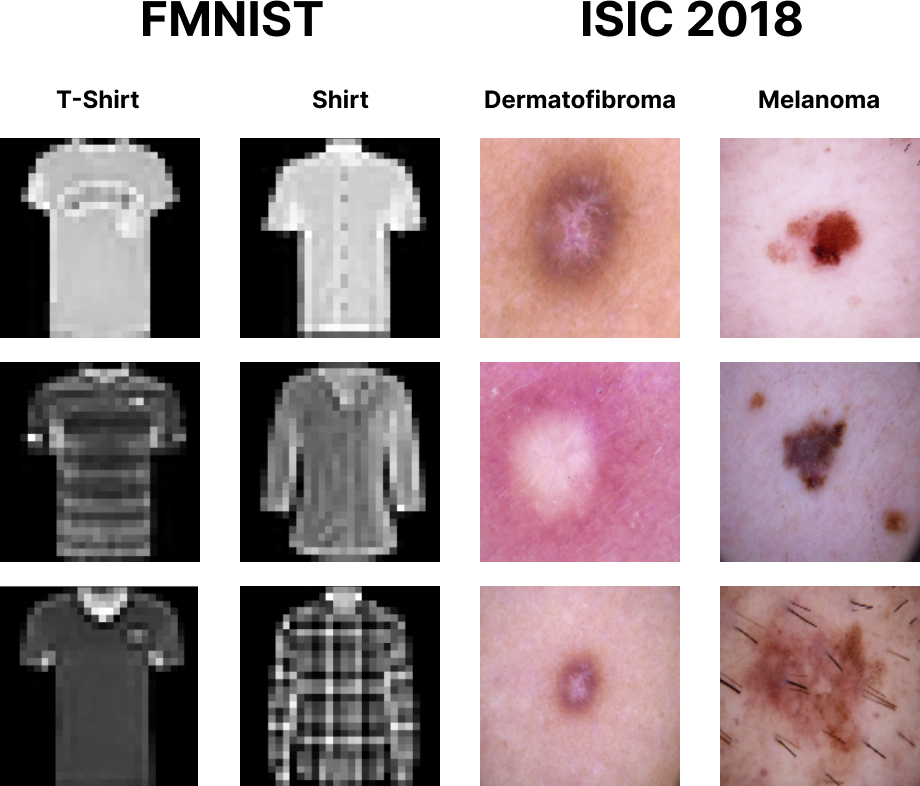}
	\caption{Sample images of the FMNIST and ISIC 2018 datasets}
	\label{fig:dataset}
\end{figure}

The FMNIST dataset consists of low-resolution ($28 \times 28$ pixels), grayscale images of ten classes of clothing. In this study, we take only two classes to form a binary classification task: the T-shirt and shirt classes. The samples are taken such that the proportion between the T-shirt and shirt images can be imbalanced depending on the scenario. On the other hand, the ISIC 2018 dataset consists of high-resolution, RGB images of seven classes of skin lesions. Following FMNIST, we use only two classes for the experiments: the melanoma and dermatofibroma classes. Illustrations of the sample images of both datasets are provided in Figure \ref{fig:dataset}.

FMNIST is chosen as a dataset since, although simple, it is a benchmark dataset to test deep learning models for computer vision. On the other hand, ISIC 2018 is chosen since it is a domain-appropriate imbalanced dataset for our model. We first apply the model (using AFCL as the loss function) to the more lightweight FMNIST dataset under various class-imbalance scenarios. This is conducted to check the appropriate values of the $\eta$ and $\gamma$ parameters of AFCL under different imbalance conditions. Afterwards, the model is applied to the ISIC 2018 dataset using the optimal parameter values obtained during the FMNIST experiments.

\subsection{Experimental Details}
The experiments are conducted using the NVIDIA Tesla P100-PCIE GPU allocated by the Google Colaboratory Pro platform. The models and loss functions are implemented using PyTorch. To process the FMNIST dataset, we use the simpler ResNet-18 architecture as the feature encoder and train it for $20$ epochs. On the other hand, to process the ISIC 2018 dataset, we use the deeper ResNet-50 as the feature encoder and train it for $40$ epochs. For both the FMNIST and ISIC 2018 datasets, the learning rate and batch size are set to $10^{-2}$ and $128$, respectively. In addition, the classification head is trained for $10$ epochs. The encoder and the classification head are both trained using the Adam optimizer. Finally, the temperature parameter $\tau$ of the contrastive loss is set to its default value of $0.07$.

The evaluation metrics utilized in the experiment are (weighted) accuracy and unweighted accuracy (UWA), both of which can be calculated from the number of true positives ($\TP$), true negatives ($\TN$), false negatives ($\FN$), and false positives ($\FP$) using the formulas
\begin{equation}
    \text{Accuracy} = \frac{\TP + \TN}{\TP + \TN + \FN + \FP}
\end{equation}
and
\begin{equation}
    \text{UWA} = \frac{1}{2}\left(\frac{\TP}{\TP+\FN} + \frac{\TN}{\TN+\FP}\right),
\end{equation}
respectively. Unlike accuracy, UWA provides the average of the individual class accuracies regardless of the number of samples in the test set of each class. UWA is an appropriate metric when the dataset is significantly imbalanced \cite{fahad}.

For heavily imbalanced datasets, a high accuracy and low UWA may mean that the model is biased towards classifying samples as part of the majority class. This indicates that the model does not properly learn from the minority samples. In contrast, a lower accuracy with high UWA indicates that the model takes significant risks to classify some samples as part of the minority class. Our aim is to construct a model that maximizes both metrics simultaneously; that is, a model that can learn unbiasedly from both the majority and minority samples with minimal misclassification error.

\subsection{Experiments using FMNIST}
The data used in the FMNIST experiment comprise of 1000 images classified as either a T-shirt or a shirt. The dataset is split 70/30 for model training and testing. The images are augmented using random rotations and random flips. We deploy 11 class-imbalance scenarios on the dataset which control the proportion between the T-shirt class and the shirt class. For example, if the the proportion is 60:40, then 600 T-shirt images and 400 shirt images are sampled to form the experimental dataset. Our proportions range from 50:50 up to 98:2.

During the first stage, the ResNet-18 encoder is trained using the AFCL. Afterwards, the classification head is trained using the CE loss during the second stage. As AFCL contains two parameters $\eta$ and $\gamma$, our goal is to tune each of these parameters independently, keeping the other parameter fixed. First, $\eta$ is tuned as we set $\gamma = 0$, followed by the tuning of $\gamma$ as we set $\eta = 0$. Each experiment is done four times in total. The average accuracy and UWA of these four runs are provided in Tables \ref{tab:eta} (for the tuning of $\eta$) and \ref{tab:gamma} (for the tuning of $\gamma$).

For the tuning of $\eta$, six values of $\eta$ are experimented on, namely $\eta \in \{0,60,120,180,240,300\}$. When $\eta = 0$, the loss function reduces to the ordinary CL. As observed in Table \ref{tab:eta}, the optimal value of $\eta$ tends to be larger when the dataset is moderately imbalanced. As the scenario goes from 60:40 to 90:10, the parameter $\eta$ that maximizes accuracy increases in value, from $\eta = 0$ when the proportion is 60:40 to $\eta = 300$ when the proportion is 90:10. In general, this indicates that the $\Ll_i^-$ term of the ACL becomes more essential to the overall loss as the dataset gets more imbalanced, confirming the reasoning contained in Subsection \ref{sec:acl}.

As seen in Table \ref{tab:gamma}, we experiment on $\gamma \in \{0,1,2,4,7,10\}$, where choosing $\gamma = 0$ means that we are using CL. Although the overall pattern of the optimal $\gamma$ is less apparent than $\eta$ of the previous experiment, some insights can still be obtained. When the scenario is between 70:30 and 90:10, the focusing parameter $\gamma$ is optimally chosen when it is larger than zero. This is in direct contrast to when the proportion is perfectly balanced (50:50), where $\gamma = 0$ is the most optimal parameter. This suggests that a larger value of $\gamma$ should be considered when class imbalance is significantly present within the dataset.

\begin{table}[t]
    \centering
    \caption{The accuracy and UWA (averaged over four independent runs) of 11 imbalance scenarios using various values of $\eta$ for the AFCL. The parameter $\gamma$ is consistently set to $0$}
    \begin{tabular}{cccccccc}
        \toprule
        \multirow{2}{*}{Scenario} & \multirow{2}{*}{Metric} & \multicolumn{6}{c}{$\eta$} \\
        \cmidrule{3-8}
        && 0 & 60 & 120 & 180 & 240 & 300 \\
        \midrule
        \multirow{2}{*}{50:50} & Accuracy & 78.92 & 77.83 & \textbf{79.75} & 71.08 & 77.17 & 78.83 \\
        & UWA & 79.00 & 78.28 & \textbf{80.32} & 72.53 & 77.87 & 79.42 \\
        \midrule
        \multirow{2}{*}{55:45} & Accuracy & \textbf{79.50} & \textbf{79.50} & 79.33 & 77.83 & 77.67 & 77.75 \\
        & UWA & 78.70 & \textbf{79.34} & 79.15 & 77.17 & 78.21 & 76.50 \\
        \midrule
        \multirow{2}{*}{60:40} & Accuracy & \textbf{84.50} & 82.92 & 82.42 & 81.33 & 82.08 & 83.17 \\
        & UWA & \textbf{83.09} & 81.82 & 81.27 & 79.71 & 81.74 & 81.66 \\
        \midrule
        \multirow{2}{*}{65:35} & Accuracy & 81.50 & \textbf{83.42} & 83.25 & 81.59 & 82.58 & 79.25 \\
        & UWA & 79.19 & \textbf{80.91} & 80.73 & 77.92 & 79.43 & 75.42 \\
        \midrule
        \multirow{2}{*}{70:30} & Accuracy & 82.50 & 84.33 & \textbf{85.08} & 82.08 & 83.42 & 83.00 \\
        & UWA & 78.41 & 78.26 & \textbf{80.91} & 77.78 & 79.14 & 75.11 \\
        \midrule
        \multirow{2}{*}{75:25} & Accuracy & 86.75 & 85.17 & 85.58 & 85.17 & \textbf{86.92} & 86.58 \\
        & UWA & 77.87 & 76.48 & 77.74 & 77.03 & \textbf{78.63} & 77.57 \\
        \midrule
        \multirow{2}{*}{80:20} & Accuracy & 86.00 & 87.25 & 87.33 & 87.92 & 87.00 & \textbf{88.25} \\
        & UWA & 76.16 & 74.65 & 76.94 & 76.28 & \textbf{77.49} & 76.97 \\
        \midrule
        \multirow{2}{*}{85:15} & Accuracy & 87.33 & 87.08 & 86.75 & 87.42 & 87.33 & \textbf{87.67} \\
        & UWA & \textbf{70.08} & 66.34 & 55.77 & 68.33 & 69.83 & 62.83 \\
        \midrule
        \multirow{2}{*}{90:10} & Accuracy & 90.83 & 91.00 & 90.83 & 90.67 & 89.50 & \textbf{91.67} \\
        & UWA & 64.91 & 68.61 & 66.11 & 64.02 & 61.77 & \textbf{72.58} \\
        \midrule
        \multirow{2}{*}{95:5} & Accuracy & \textbf{94.42} & 93.33 & 93.42 & 94.00 & 92.83 & 93.25 \\
        & UWA & 54.77 & \textbf{60.70} & 54.24 & 50.00 & 49.38 & 54.80 \\
        \midrule
        \multirow{2}{*}{98:2} & Accuracy & 97.42 & 97.83 & 98.08 & 98.08 & \textbf{98.33} & 98.08 \\
        & UWA & 52.45 & 52.66 & \textbf{55.87} & \textbf{55.87} & 49.83 & 52.79 \\
        \bottomrule
    \end{tabular}
    \label{tab:eta}
\end{table}

\begin{table}[t]
    \centering
    \caption{The accuracy and UWA (averaged over four independent runs) of 11 imbalance scenarios using various values of $\gamma$ for the AFCL. The parameter $\eta$ is consistently set to $0$}
    \begin{tabular}{cccccccc}
        \toprule
        \multirow{2}{*}{Scenario} & \multirow{2}{*}{Metric} & \multicolumn{6}{c}{$\gamma$} \\
        \cmidrule{3-8}
        && 0 & 1 & 2 & 4 & 7 & 10 \\
        \midrule
        \multirow{2}{*}{50:50} & Accuracy & \textbf{78.08} & 74.83 & 77.08 & 77.58 & 76.58 & 77.50 \\
        & UWA & \textbf{77.70} & 74.84 & 76.77 & 77.55 & 76.55 & 77.25\\
        \midrule
        \multirow{2}{*}{55:45} & Accuracy & 80.17 & 81.25 & 80.75 & 80.00 & \textbf{81.75} & 76.83 \\
        & UWA & 80.14 & 81.19 & 80.69 & 79.96 & \textbf{81.70} & 76.82\\
        \midrule
        \multirow{2}{*}{60:40} & Accuracy & 79.42 & 78.50 & 77.92 & 80.17 & \textbf{80.67} & 80.08 \\
        & UWA & \textbf{84.42} & 83.42 & 80.00 & 83.00 & 82.42 & 82.92\\
        \midrule
        \multirow{2}{*}{65:35} & Accuracy & \textbf{84.42} & 83.42 & 80.00 & 83.00 & 82.42 & 82.92 \\
        & UWA & \textbf{81.98} & 81.22 & 77.87 & 80.39 & 80.68 & 80.16\\
        \midrule
        \multirow{2}{*}{70:30} & Accuracy & 83.75 & 83.83 & 82.17 & 82.58 & \textbf{84.83} & 82.25 \\
        & UWA & 79.64 & 79.18 & 77.82 & 77.51 & \textbf{79.67} & 78.71\\
        \midrule
        \multirow{2}{*}{75:25} & Accuracy & 85.42 & \textbf{86.17} & 84.42 & 84.83 & 85.75 & 86.00 \\
        & UWA & 76.27 & \textbf{79.85} & 77.08 & 76.41 & 77.34 & 78.47\\
        \midrule
        \multirow{2}{*}{80:20} & Accuracy & 89.33 & \textbf{89.58} & 87.67 & 89.42 & 87.33 & 88.00 \\
        & UWA & 77.59 & 78.67 & 78.43 & \textbf{79.31} & 78.97 & 70.12\\
        \midrule
        \multirow{2}{*}{85:15} & Accuracy & 87.42 & 89.00 & 88.17 & 88.33 & 89.08 & \textbf{90.08} \\
        & UWA & 64.97 & 72.08 & 71.99 & 71.47 & 71.95 & \textbf{77.04} \\
        \midrule
        \multirow{2}{*}{90:10} & Accuracy & 92.42 & 92.33 & \textbf{93.42} & 93.25 & 92.58 & 91.25 \\
        & UWA & 64.00 & 67.94 & 66.04 & 74.42 & \textbf{80.54} & 68.35\\
        \midrule
        \multirow{2}{*}{95:5} & Accuracy & 94.17 & 93.17 & \textbf{95.33} & 95.00 & 94.00 & 95.09 \\
        & UWA & \textbf{62.13} & 53.11 & 57.64 & 59.17 & 55.22 & 55.82\\
        \midrule
        \multirow{2}{*}{98:2} & Accuracy & \textbf{96.92} & \textbf{96.92} & 95.00 & 96.00 & \textbf{96.92} & 96.67 \\
        & UWA & \textbf{56.59} & 51.56 & 55.61 & 52.63 & 53.10 & 52.98\\
        \bottomrule
    \end{tabular}
    \label{tab:gamma}
\end{table}

\subsection{Experiments using ISIC 2018}
From the ISIC 2018 dataset, a total of 1113 melanoma images and 115 dermatofibroma images are combined to create the experimental dataset. As with the previous experiment, the dataset is split 70/30 for training and testing. The images are resized to $128 \times 128$ pixels. The ResNet-50 encoder is trained using one of the available contrastive losses, which include CL/FCL as baselines and ACL/AFCL as the proposed loss functions. The classification head is trained using FL as the loss function with its focusing parameter set to $\gamma = 2$.

The proportion between the melanoma class and the dermatofibroma class in the experimental dataset is close to 90:10. Using results from Tables \ref{tab:eta} and \ref{tab:gamma} as a heuristic for determining the optimal parameter values, we set $\eta = 300$ and $\gamma = 2, 7$. It is worth mentioning that even though $\gamma = 2$ produces the best accuracy in the FMNIST experiment, the UWA of the resulting model is quite poor. However, we decide to include this value in this experiment for completeness.

The results of this experiment is given in Table \ref{tab:isic}. As in the previous section, each experiment is conducted four times, so the table lists the average accuracy and UWA of these four runs for each contrastive loss tested. Each run, which includes both model training and testing, is completed in roughly 80 minutes using our computational setup.

From Table \ref{tab:isic}, CL and ACL performs the worst in terms of UWA and accuracy, respectively. However, ACL gives the best UWA among all losses. This may indicate that ACL encourages the model to take the risky approach of classifying some samples as part of the minority class at the expense of accuracy. Overall, AFCL with $\eta = 300$ and $\gamma = 7$ emerges as the best loss in this experiment, producing the best accuracy and the second-best UWA behind ACL. This leads us to conclude that the AFCL, with optimal hyperparameters chosen, is superior to the vanilla CL and FCL.

\begin{table}[t]
    \centering
    \caption{The accuracy and UWA (averaged over four independent runs) of the model when trained using various contrastive losses}
    \begin{tabular}{ccc}
        \toprule
        Loss function & Accuracy & UWA \\
        \midrule
        CL \cite{khosla} & 93.00 &  72.25 \\
        FCL \cite{zhang_fcl} & 93.07 & 74.34 \\
        ACL ($\eta$ = 300) & 85.94 & \textbf{75.54} \\
        AFCL ($\eta$ = 300, $\gamma$ = 2) & 92.39 & 74.36 \\
        AFCL ($\eta$ = 300, $\gamma$ = 7) & \textbf{93.75} & 74.62 \\
        \bottomrule
    \end{tabular}
    \label{tab:isic}
\end{table}

\section{Conclusion and Future Work}
In this work, we introduced an asymmetric version of both contrastive loss (CL) and focal contrastive loss (FCL) referred to as ACL and AFCL, respectively. These asymmetric variants of the contrastive loss were proposed to provide more focus on the minority class. The experimental model used was a two-stage architecture consisting of a feature learning stage and a classifier fine-tuning stage. This model was applied to the FMNIST and ISIC 2018 imbalanced datasets using various contrastive losses. Our results show that AFCL was able to outperform CL and FCL in terms of both weighted and unweighted accuracies. On the ISIC 2018 binary classification task, AFCL, with $\eta = 300$ and $\gamma = 7$ as hyperparameters, achieved an accuracy of 93.75\% and an unweighted accuracy of 74.62\%. This is in contrast to FCL, which achieved 93.07\% and 74.34\% on both metrics, respectively.

The experiments of this research were conducted using datasets consisting of approximately 1000 total images. In the future, the experimental model may be applied to larger-scale datasets in order to test its scalability. In addition, other models based on ACL and AFCL can also be developed for specific datasets, preferably within the realm of multiclass classification.

\printbibliography

\renewcommand\thesection{\Alph{section}}
\setcounter{section}{0}

\section{Axioms for Entropy}\label{app:entropy}
In his landmark paper, Shannon \cite{shannon} introduced the notion of \textit{entropy} $H(X)$ of a random variable $X$. Entropy measures the amount of information contained in $X$, usually in bits. For example, a fair coin toss contains one bit of information; the $0$ bit can represent the heads whereas the $1$ bit can represent the tails. On the other hand, an unfair coin toss whose coin always lands on heads gives no meaningful information. Hence, the trial can be conveyed using zero bits.

This section aims to construct the theory of entropy via an axiomatic approach. First, a collection of axioms, known as the \textit{Shannon--Khinchin axioms} \cite{khinchin}, is employed to give desired properties of the function $H(\cdot)$. Then, it is shown that the usual formula for $H(X)$ follows uniquely from these axioms. The presentation of the axioms in this section follows a set of notes provided by Gowers \cite{gowers}.

Suppose that $X$ and $Y$ are discrete random variables taking values in finite spaces $\X$ and $\Y$, respectively. Let $p_x = P_X(x) = \Pp[X = x]$ and $q_y = P_Y(y) = \Pp[Y = y]$ for $x \in \X$ and $y \in \Y$. The first axiom is motivated using the coin toss example. Since a fair coin toss is expected to contain one bit of information, the following axiom is obtained.

\begin{axiom}[Normalization]\label{ax:norm}
If $\lvert \X \rvert = 2$ and $X$ has a uniform distribution, then $H(X) = 1$.
\end{axiom}

Also, $H(X)$ depends only on the probability distribution of $X$. Consequently, if $Y$ is another random variable that has an identical distribution to $X$, then $H(Y) = H(X)$.

\begin{axiom}[Invariance]\label{ax:inv}
$H(X)$ depends only on the probability distribution of $X$, and not on any other factor.
\end{axiom}

Going back to the coin toss example, we would like to ensure that a coin toss contains the most information when it is fair. In general, the following axiom is assumed.

\begin{axiom}[Maximality]\label{ax:max}
Assuming $\lvert \X \rvert$ is fixed, $H(X)$ is maximized when $X$ is uniform.
\end{axiom}

In addition, the value of $H(X)$ should not increase when impossible samples are added to $\X$.

\begin{axiom}[Extensibility]\label{ax:ext}
If $\X \subset \Y$ with $p_x = q_x$ for every $x \in \X$ (and thus $q_y = 0$ for every $y \in \Y \setminus \X$), then $H(Y) = H(X)$.
\end{axiom}

To state the next axiom, two notions on entropy are first introduced. The \textit{joint entropy} $H(X, Y)$ is simply the entropy of the joint random variable $(X,Y)$, and the \textit{conditional entropy} $H(Y \mid X)$ is defined as
\begin{equation}
    H(Y \mid X) = \sum_{x \in \X} p_x H(Y \mid X = x).
\end{equation}
Conditional entropy measures the average amount of information contained in $Y$ given the value of $X$.

\begin{axiom}[Additivity]\label{ax:add}
$H(X,Y) = H(X) + H(Y \mid X)$.
\end{axiom}

If $X$ and $Y$ are independent, then $H(Y \mid X) = H(Y)$. Therefore, $H(X,Y) = H(X) + H(Y)$ in that case. In general, if $X_1, \dots, X_n$ are independent, then $H(X_1, \dots, X_n) = \sum_{i=1}^n H(X_i)$.

Suppose that $\X = \{1, \dots, n\}$. Since $H(X)$ only depends on the distribution of $X$ by Axiom \ref{ax:inv}, the function $H(X)$ can instead be seen as a function $H(p_1, \dots, p_n)$. The next axiom states that $H(p_1, \dots, p_n)$ is continuous on the space
\begin{equation}
    S = \{(p_1, \dots, p_n) \in [0,1]^n \mid p_1 + \dots + p_n = 1\}.
\end{equation}

\begin{axiom}[Continuity]\label{ax:cont}
$H(X)$ is continuous with respect to all probabilities $p_x$.
\end{axiom}

From Axioms \ref{ax:norm}--\ref{ax:cont}, the formula for $H(X)$ is uniquely determined as shown in the following theorem.

\begin{theorem_app}\label{thm:entropy}
Let $H(X)$ be a function defined for any discrete random variable $X$ that takes values in a finite set $\X$. This function satisfies Axioms \ref{ax:norm}--\ref{ax:cont} if and only if
\begin{equation}\label{eq:entropy}
    H(X) = - \sum_{x \in \X} p_x \log(p_x),
\end{equation}
where the logarithm is to the base $2$ and we set $0 \cdot \log(0) = 0$.
\end{theorem_app}

\begin{figure}[t]
	\centering
	\includegraphics[width=0.6\textwidth]{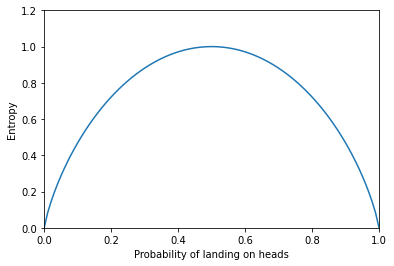}
	\caption{A graph illustrating the entropy of a coin toss with varying fairness}
	\label{fig:entropy}
\end{figure}

The proof of \ref{thm:entropy} is provided in Appendix \ref{app:entropy_proof}. Looking back at the coin toss example, Figure \ref{fig:entropy} illustrates the graph of $H(X)$ when $X$ is either heads or tails with probabilities $p$ and $1-p$, respectively. Entropy is maximized when the coin is fair, and it decreases in a continuous manner to zero as the coin becomes less fair.

The formula for entropy in Eq.\ (\ref{eq:entropy}) can be expressed in the form of an expectation:
\begin{equation}
    H(X) = \E_{P_X}\left[-\log(P_X(X))\right].
\end{equation}
Likewise, joint entropy and conditional entropy can be expressed as
\begin{equation}
    H(X,Y) = \E_{P_{(X,Y)}}\left[-\log(P_{(X,Y)}(X,Y))\right]
\end{equation}
and
\begin{align*}
    H(Y \mid X) &= \sum_{x \in \X} p_x H(Y \mid X = x)\\
    &= - \sum_{x \in \X} \left(p_x \sum_{y \in \Y} P_{Y \mid X}(y \mid x) \log\left(P_{Y \mid X}(y \mid x)\right)\right)\\
    &= - \sum_{x \in \X} \sum_{y \in \Y} p_x P_{Y \mid X}(y \mid x) \log\left(P_{Y \mid X}(y \mid x)\right)\\
    &= - \sum_{(y,x)} P_{(Y, X)}(y, x) \log\left(P_{Y \mid X}(y \mid x)\right)\\
    &= \E_{P_{(Y,X)}}\left[-\log\left(P_{Y \mid X}(Y \mid X)\right)\right].
\end{align*}

\section{Proof of Theorem \ref{thm:entropy}}\label{app:entropy_proof}
The arguments used in this proof are adapted from \cite{gowers, khinchin}. We first verify one direction of Theorem \ref{thm:entropy}.

\begin{lemma_app}\label{lem:entropy}
The formula for $H(X)$ given in Eq.\ (\ref{eq:entropy}) satisfies Axioms \ref{ax:norm}--\ref{ax:cont}.
\end{lemma_app}

\begin{proof}
It is trivial to show that the normalization, invariance, extensibility, and continuity axioms hold, so we focus on proving the maximality and additivity axioms.

For maximality, we need to utilize Jensen's inequality \cite{ajjanagadde} applied on the concave function $\log$. This inequality takes the form
\begin{equation}\label{eq:jensen}
    \E[\log Y] \le \log(\E[Y]).
\end{equation}
For any random variable $X$,
\begin{align*}
    H(X) &= \E\left[\log\left(\frac{1}{P(X)}\right)\right]\\
    &\le \log\left(\E\left[\frac{1}{P(X)}\right]\right) &\left(\text{by Eq.\ (\ref{eq:jensen}), where } Y = \frac{1}{P(X)}\right)\\
    &= \log\left(\sum_{x \in \X} p_x \cdot \frac{1}{p_x}\right)\\
    &= \log(\lvert \X \rvert).
\end{align*}
Since $\log(\lvert \X \rvert)$ is the entropy of a uniform random variable on $\X$, the entropy $H(X)$ is maximized $X$ is uniform.

For additivity, we need to prove that $H(X,Y) = H(X) + H(Y \mid X)$. Writing $p_{xy} = P_{(X,Y)}(x,y)$, we have
\begin{align*}
    H(X,Y) &= - \sum_{x \in \X} \sum_{y \in \Y} p_{x,y} \log(p_{x,y})\\
    &= - \sum_{x \in \X} \sum_{y \in \Y} p_{x,y} \log\left(p_xP_{Y \mid X}(y \mid x)\right)\\
    &= - \sum_{x \in \X} \sum_{y \in \Y} p_{x,y} \left(\log(p_x) + \log\left(P_{Y \mid X}(y \mid x)\right)\right)\\
    &= - \sum_{x \in \X} \sum_{y \in \Y} p_{x,y}\log(p_x) - \sum_{x \in \X} \sum_{y \in \Y} p_{x,y}\log\left(P_{Y \mid X}(y \mid x)\right).
\end{align*}
We can obtain
\begin{equation*}
    - \sum_{x \in \X} \sum_{y \in \Y} p_{x,y}\log(p_x) = - \sum_{x \in \X} p_x\log(p_x) = H(X)
\end{equation*}
and
\begin{align*}
    - \sum_{x \in \X} \sum_{y \in \Y} p_{x,y}\log\left(P_{Y \mid X}(y \mid x)\right) &= - \sum_{x \in \X} \sum_{y \in \Y} p_x P_{Y \mid X}(y \mid x)\log\left(P_{Y \mid X}(y \mid x)\right)\\
    &= - \sum_{x \in \X} \left(p_x \sum_{y \in \Y} P_{Y \mid X}(y \mid x)\log\left(P_{Y \mid X}(y \mid x)\right)\right)\\
    &= \sum_{x \in \X} p_x\ H(Y \mid X = x)\\
    &= H(Y \mid X).
\end{align*}
Therefore, $H(X,Y) = H(X) + H(Y \mid X)$.
\end{proof}

To ease the notation, we can assume that $\X = \{1, \dots, n\}$ and write $H(p_1, \dots, p_n)$ in place of $H(X)$ by the invariance axiom. For brevity, $L(n)$ is defined as the entropy of a uniform random variable with $\lvert \X \rvert = n$. In other words,
\begin{equation}
    L(n) = H\left(\frac{1}{n}, \dots, \frac{1}{n}\right).
\end{equation}

\begin{lemma_app}\label{lem:L(n)}
The following properties hold for the function $L(n)$:
\begin{enumerate}
    \item $L(n)$ is non-decreasing.
    \item $L(n^m) = m L(n)$.
    \item $L(2^k) = k$.
    \item $L(n) = \log(n)$.
\end{enumerate}
\end{lemma_app}

\begin{proof}
1. For every natural number $n$,
\begin{align*}
    L(n) &= H\left(\frac{1}{n}, \dots, \frac{1}{n}, 0\right) &(\text{by the extensibility axiom})\\
    &\le H\left(\frac{1}{n+1}, \dots, \frac{1}{n+1}\right) &(\text{by the maximality axiom})\\
    &= L(n+1).
\end{align*}
Since $n$ is arbitrary, this proves that $L(n)$ is a non-decreasing function of $n$.

2. Let $X$ be a uniform random variable on $\X$ with $\lvert\X\rvert = n$. Then $H(X) = L(n)$. Now let $X_1, \dots, X_m$ be i.i.d.\ random variables with distribution identical to $X$. Since the joint variable $(X_1, \dots, X_m)$ is uniform, we obtain
\begin{equation*}
    L(n^m) = H(X_1, \dots, X_m) = \sum_{i=1}^m H(X_i) = mH(X) = mL(n).
\end{equation*}

3. By the normalization axiom, $L(2) = 1$. Therefore, $L(2^k) = k\cdot L(2) = k$.

4. We aim to prove that $\lvert L(n)-\log(n) \rvert \le 1/m$ for every $n, m \ge 1$. Fix $n$ and $m$. Now let $k$ be the unique integer such that the inequality
\begin{equation}\label{eq:ineq_1}
    2^k \le n^m \le 2^{k+1}
\end{equation}
holds. Applying the non-decreasing function $L(\cdot)$ on (\ref{eq:ineq_1}), we obtain
\begin{equation}\label{eq:ineq_2}
    k \le m L(n) \le k+1.
\end{equation}
Applying the non-decreasing function $\log(\cdot)$ on (\ref{eq:ineq_1}), we also obtain
\begin{equation}\label{eq:ineq_3}
    k \le m \log(n) \le k+1.
\end{equation}
Both (\ref{eq:ineq_2}) and (\ref{eq:ineq_3}) imply that
\begin{equation}\label{eq:ineq_4}
    \lvert m L(n) - m \log (n)\rvert \le (k+1) - k = 1.
\end{equation}
We can obtain $\lvert L(n) - \log (n)\rvert \le 1/m$ by dividing both sides of (\ref{eq:ineq_4}) by $m$. As a consequence, $L(n) = \log(n)$.
\end{proof}

The fourth property of Lemma \ref{lem:L(n)} infers that Theorem \ref{thm:entropy} holds for uniform random variables $X$. We are now ready to complete the proof of Theorem \ref{thm:entropy} for any random variable $X$.

\begin{proof}[Proof of Theorem \ref{thm:entropy}]
One half of Theorem \ref{thm:entropy} is proved in Lemma \ref{lem:entropy}. It remains to prove that a function $H(X) = H(p_1, \dots, p_n)$ which satisfies Axioms \ref{ax:norm}--\ref{ax:cont} is necessarily equal to $- \sum_{i = 1}^n p_i \log(p_i)$. Without loss of generality, we can assume that $p_1, \dots, p_n$ are all rational. Indeed, since the rationals are dense in the reals, the theorem would still hold for real values $p_1, \dots, p_n$ by the continuity axiom (Axiom \ref{ax:cont}).

Let $p_i = g_i/g$ for $1 \le i \le n$, where each $g_i$ is a positive integer and $\sum_{i=1}^n g_i = g$. Define a random variable $Y$ dependent on $X$ such that $\lvert \Y \rvert = g$ and $\Y$ is partitioned into $n$ disjoint groups $\Y_1, \dots, \Y_n$ containing $g_1, \dots, g_n$ values, respectively. If it is given that $X = i$, where $1 \le i \le n$, then all the values in $\Y_i$ have the same probability $1/g_i$, and values from other groups have probability zero. It follows that
\begin{align*}
    H(Y \mid X) &= \sum_{i = 1}^n p_i H(Y \mid X = i)\\
    &= \sum_{i = 1}^n p_i L(g_i)\\
    &= \sum_{i = 1}^n p_i \log(g\cdot p_i)\\
    &= \sum_{i = 1}^n p_i (\log(g) + \log(p_i))\\
    &= \log(g) + \sum_{i = 1}^n p_i \log(p_i).
\end{align*}
In addition, $(X,Y)$ and $Y$ are identically distributed since $X$ is completely dependent on $Y$. The joint variable $(X,Y)$ thus has a total of $g$ possible values, and each value has the same probability $1/g$ of occurring. By the additivity axiom,
\begin{align*}
    H(X) &= H(X,Y) - H(Y \mid X)\\
    &= L(g) - \log(g) - \sum_{i = 1}^n p_i \log(p_i)\\
    &= - \sum_{i = 1}^n p_i \log(p_i).
\end{align*}
This proves that $H(p_1, \dots, p_n) = - \sum_{i = 1}^n p_i \log(p_i)$ for rational $p_1, \dots, p_n$. The full statement of the theorem follows from continuity, as explained at the beginning of the proof.
\end{proof}

\end{document}